\documentclass[10pt,twocolumn]{ICCAS}

\usepackage{diagbox}
\usepackage{amsmath}
\usepackage{amssymb}
\usepackage{graphicx}
\usepackage{float}
\usepackage{caption}
\usepackage{subcaption}
\usepackage[noadjust]{cite}

\begin{document}


\title{Reflection-Robust Stereo Visual-Inertial Odometry via \\ Kalman-Filtered Ground-Height Estimation}

\author{Uihyun Hwang${}^{1}$, Sungjae Shin${}^{1}$, Dongjae Kim${}^{1}$, and Hyun Myung${}^{1*}$}


\affils{ ${}^{1}$School of Electrical Engineering, Korea Advanced Institute of Science and Technology, Daejeon, 34141, South Korea ({joshuahwang, pootti, zeze0214, hmyung}@kaist.ac.kr)  \\ {\small${}^{*}$ Corresponding author}}

\abstract{Visual-inertial odometry relies on visual features that correspond to 3D points with consistent projections across sequential frames. In reflective indoor-floor environments, however, specular floor patterns can be detected and triangulated as virtual features below the physical ground plane, producing unreliable feature associations and inconsistent visual residuals. This paper proposes a lightweight reflection-aware front-end for feature-based stereo VIO. Rather than suppressing all floor-region features, the proposed method estimates the ground height in real time and rejects only feature candidates whose inferred 3D positions are below the estimated ground height. A floor-region mask restricts the validation process to reflection-prone regions, while stereo height uncertainty and pose uncertainty are used to construct frame-level ground-height measurements. A scalar Kalman filter with Normalized Innovation Squared (NIS) gating then provides a stable ground-height estimate, and an uncertainty-aware safety margin is used for selective below-ground virtual feature rejection. Experiments on real reflective-floor indoor sequences show that the proposed front-end generally improves trajectory accuracy in reflection-dominant environments while maintaining accurate ground-height estimates and recovering from actual ground-height changes.}



\maketitle



\section{INTRODUCTION}
Visual-inertial odometry (VIO) estimates camera motion by combining inertial measurements with visual feature observations. In feature-based stereo-inertial systems, feature detection, tracking, stereo matching, and triangulation precede the formation of visual constraints~\cite{vinsmono,qin2019a,dynavins,dynavinspp,uvslam,mirvio}. The reliability of these front-end features is therefore critical, as erroneous feature geometry degrades residual consistency and odometry stability.

Indoor mobile robots frequently operate on reflective floors such as polished tiles or coated concrete, where ceiling lights, window illumination, and surrounding structures are reflected and produce high-contrast image patterns. Standard feature extractors~\cite{shi1994good,orb2011} treat these reflected patterns as valid features, even though they do not correspond to stable 3D structures. The severity of this effect depends on the geometry of the surface. Herbert et al.~\cite{herbert2024mirror} showed that geometrically consistent virtual features from smooth mirrors can occasionally provide useful odometry constraints, whereas Pyykola et al.~\cite{pyykola2024nonlambertian} demonstrated that non-Lambertian surfaces such as tiles or windows yield view-dependent appearances that cause false matches, erroneous depth, and trajectory drift. Indoor reflective floors fall into the latter case, producing view-dependent specular patterns that violate multi-view consistency and degrade feature-based VIO.

\begin{figure}[!t]
\centering
\includegraphics[width=\columnwidth]{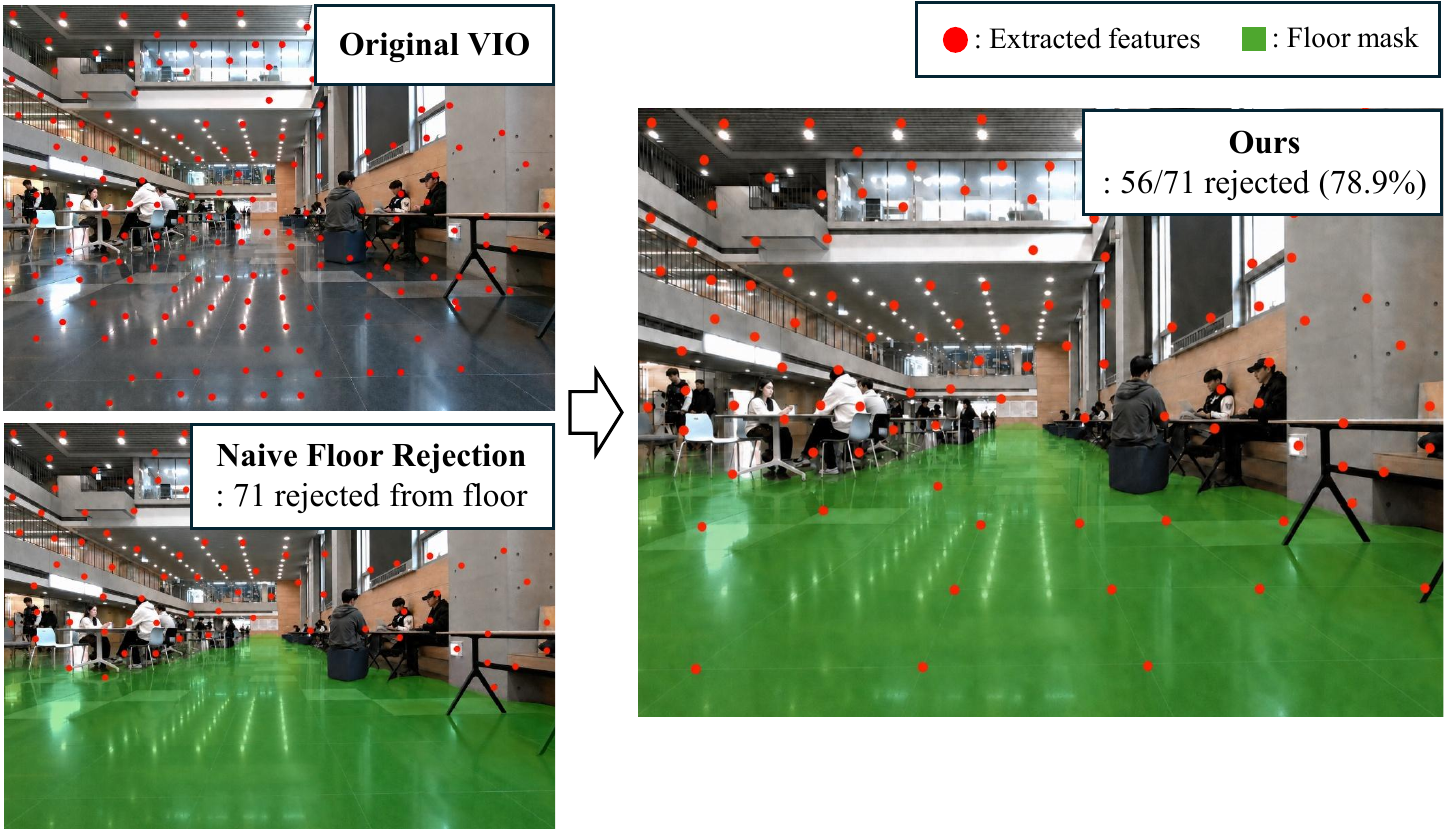}
\vspace{-3.7ex}
\caption{Comparison of feature extraction results on a reflective floor in the Yonsei environment. Of the 71 features extracted within the floor-mask region, ours rejects 56 (78.9\%) below-ground virtual candidates while preserving valid on-floor features.}
\label{fig:reflective_floor_features}
\end{figure}

As illustrated in Fig.~\ref{fig:reflective_floor_features}, the original VIO front-end extracts many features from the floor region, including reflection-induced virtual features. A naive floor rejection strategy that simply discards all features in the floor region could suppress these reflections. However, it also removes physically valid on-floor features, such as tile edges or surface texture, which can still provide meaningful visual information for VIO. This naive rejection strategy is overly conservative and may degrade the front-end, especially when background features are sparse. 

\begin{figure*}[t]
\begin{center}
\includegraphics[width=0.98\textwidth]{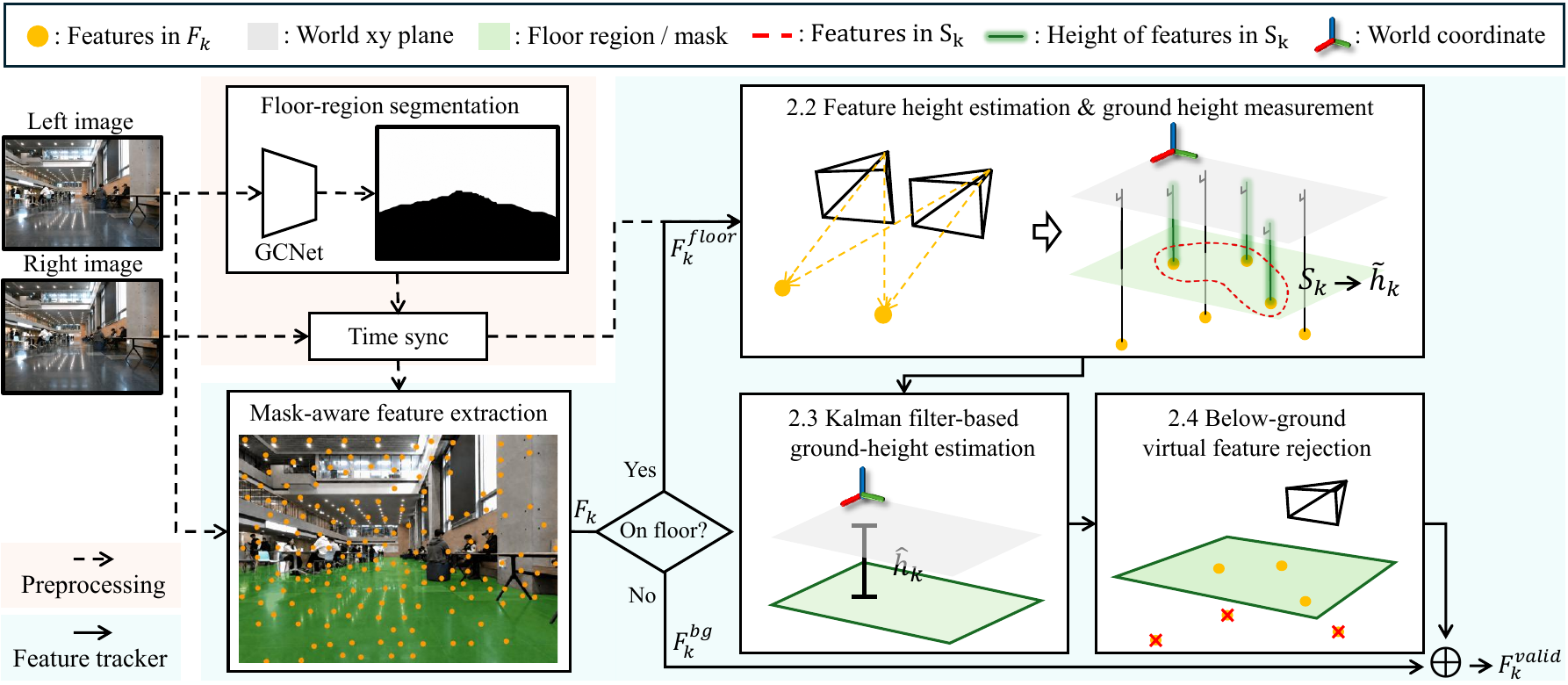}
\caption{Overview of the proposed reflection-robust VIO front-end pipeline. A GCNet~\cite{yang2025gcnet} mask splits $F_k$ into floor-region features $F_k^{\mathrm{floor}}$ and background features $F_k^{\mathrm{bg}}$. Only $F_k^{\mathrm{floor}}$ drives $\tilde{h}_k$, the Kalman-filtered estimate $\hat{h}_k$, and below-ground virtual feature rejection; validated floor and background features are merged into $F_k^{\mathrm{valid}}$ for the VIO estimator.}
\label{fig:pipeline}
\end{center}
\end{figure*}

A natural way to realize such a geometry-based criterion is to estimate the ground plane explicitly and reject features that lie far from it. Zhou et al.~\cite{zhou2025rgbd} followed this idea in an RGB-D SLAM system, fitting the ground plane from dense depth via RANSAC and discarding features distant from the fitted floor. However, per-pixel depth processing with iterative RANSAC is computationally expensive for a lightweight stereo VIO front-end. Moreover, when virtual features dominate the floor observations, the RANSAC inliers become contaminated and the fitted plane is biased below the true ground, which undermines the very geometric reference used for rejection.

This paper proposes a reflection-robust stereo VIO front-end for selective below-ground virtual feature rejection. Instead of fitting a dense ground plane, the proposed method maintains an adaptive scalar ground-height estimate from sparse stereo features and Kalman filtering~\cite{welch1995kalman}, and rejects features whose triangulated positions fall below the estimated ground height by more than an uncertainty-aware margin. This keeps the rejection criterion robust to reflection-induced contamination while retaining real-time operation. The contributions of this paper are summarized below.

1) We propose a lightweight Kalman filter-based ground-height estimation method that adapts to robot pose variation and physical ground-height changes.

2) We present a selective below-ground virtual feature rejection strategy for VIO front-end reliability, which is applicable to feature-based stereo VIO pipelines.


\section{PROPOSED METHOD}
As shown in Fig.~\ref{fig:pipeline}, the proposed front-end partitions the stereo-matched feature set $F_k$ into floor-region features $F_k^{\mathrm{floor}}$ and background features $F_k^{\mathrm{bg}}$. Features in $F_k^{\mathrm{floor}}$ form a ground-height measurement $\tilde{h}_k$, which is refined into $\hat{h}_k$ by a scalar Kalman filter with NIS validation. Features inconsistent with $\hat{h}_k$ are rejected, and the remaining floor features are merged with $F_k^{\mathrm{bg}}$ to form $F_k^{\mathrm{valid}}$. Each stage is detailed below.

\subsection{Floor-Region Segmentation and Mask-Aware Feature Extraction}
A 2D floor mask is provided by a Golden Cudgel Network (GCNet) real-time segmentation model~\cite{yang2025gcnet}. The floor mask restricts geometric validation to reflection-prone regions. After temporal tracking and stereo matching, the feature set $F_k$ is partitioned into floor-region features $F_k^{\mathrm{floor}}$ and background features $F_k^{\mathrm{bg}} = F_k \setminus F_k^{\mathrm{floor}}$, according to the left-image mask. The former includes both reflection-induced virtual and valid on-floor features and undergoes the proposed validation, while $F_k^{\mathrm{bg}}$ directly bypasses the filter and is retained as valid input.

\subsection{Feature Height Estimation and Ground-Height Measurement}
This stage converts features in $F_k^{\mathrm{floor}}$ into a single per-frame ground-height measurement $\tilde{h}_k$ and its uncertainty $U_k$, which serve as the observation input for the Kalman filter in Section~2.3. To this end, we first compute each feature's height along gravity, construct a support set of features for the current-frame measurement, and aggregate their heights into $\tilde{h}_k$ as a weighted sum, with $U_k$ propagated from stereo and pose uncertainty.

\textbf{Estimating feature height along gravity.}
For a rectified stereo pair, feature $i$ is described by the stereo measurement and disparity
\begin{equation}
m_i = \left[ u_{L,i}, v_{L,i}, u_{R,i} \right]^T, \quad
d_i = u_{L,i} - u_{R,i}.
\label{eq:stereo_measurement}
\end{equation}
Let $f$ denote the focal length, $(c_x, c_y)$ the principal point, and $B$ the stereo baseline. The corresponding camera-frame 3D point is obtained by triangulation:
\begin{equation}
p_i^c =
\left[
\dfrac{(u_{L,i}-c_x)B}{d_i},\,
\dfrac{(v_{L,i}-c_y)B}{d_i},\,
\dfrac{f B}{d_i}
\right]^T.
\label{eq:triangulation}
\end{equation}
Let $(R_{wc}, t_{wc})$ be the camera-to-world pose from the current VIO estimate. The point $p_i^c$ is transformed to the world frame and projected onto the gravity vector:
\begin{equation}
p_i^w = R_{wc}\, p_i^c + t_{wc}, \quad
z_i = (g^w)^T p_i^w,
\label{eq:height_projection}
\end{equation}
$g^w$ is the unit direction of gravity in the world frame, fixed after VIO initialization, and $z_i$ is the coordinate of feature $i$ along $g^w$; it increases downward since $g^w$ points downwards in the world coordinate. 

\textbf{Support set construction.} The support set $S_k \subset F_k^{\mathrm{floor}}$ is defined as a fixed number $N$ of features used to form the ground-height measurement. Candidates near the mask boundary are first discarded, since segmentation errors often occur near the boundary. Among the rest,
$N$ features with the smallest $z_i$, i.e. those with the least downward displacement along gravity, are selected as $S_k$. However, since on-floor and below-ground virtual points cannot be separated by height alone, $S_k$ may remain contaminated. The effect of such contamination is mitigated downstream through uncertainty-based measurement weighting and NIS gating.

\textbf{Forming the per-frame measurement.}
The ground-height measurement at frame $k$ is formed as the weighted average of the support-feature heights,
\begin{equation}
\tilde{h}_k = \sum_{i \in S_k} w_i z_i,
\quad \text{s.t.} \quad
\sum_{i\in S_k} w_i = 1,\; w_i \geq 0,
\label{eq:weighted_measurement}
\end{equation}
where the weights $w_i$ are chosen to minimize the measurement uncertainty $U_k$.

\textbf{Modeling the measurement uncertainty.}
Each $z_i$ is built from stereo triangulation and the current VIO pose (Eq.~\ref{eq:height_projection}), so the uncertainty of $\tilde{h}_k$ has two origins: pixel-level stereo noise and estimator-level pose error. Since the pose error further splits into translation and rotation error, a first-order analysis decomposes $U_k$ into a stereo term, a translation term, and a rotation term.

\textit{Stereo term.} Following~\cite{wong2017uncertainty}, we model the pixel localization uncertainty of feature $i$ as an anisotropic covariance $\Sigma_{L,i}$ from the local image structure tensor and propagate it to the gravity-aligned height. Under rectification, only $u_R$ affects disparity, so the right view adds a scalar $\sigma_{R,u,i}^2$. With triangulation Jacobian $J_i = \partial p_i^c / \partial m_i$ and height sensitivity $b_i = J_i^T g^c$, the stereo measurement covariance $\Sigma_{m,i} = \mathrm{diag}(\Sigma_{L,i},\, \sigma_{R,u,i}^2)$ gives the per-feature stereo-induced height uncertainty
\begin{equation}
U_i^{\mathrm{stereo}} = b_i^T \Sigma_{m,i}\, b_i .
\label{eq:stereo_uncertainty}
\end{equation}

\textit{Translation term.} A translational pose error shifts the height of all support features in common. Projecting the translational covariance $\Sigma_t$ onto the gravity axis gives
\begin{equation}
U_t = (g^w)^T \Sigma_t\, g^w .
\label{eq:translation_uncertainty}
\end{equation}
This term is added once rather than per feature, as it is a shared systematic error. In the implementation $\Sigma_t$ is approximated as $\sigma_t^2 I$ with an empirically determined $\sigma_t$.

\textit{Rotation term.} A small attitude perturbation $\delta\theta$ induces a per-feature height perturbation $\delta z_{i,\theta} \approx q_i^T \delta\theta$, where
\begin{equation}
q_i = p_i^c \times g^c .
\label{eq:rotation_sensitivity}
\end{equation}
The attitude error is approximated as isotropic, $\mathrm{Cov}(\delta\theta)\approx\sigma_\theta^2 I$. Since all support features share the same $\delta\theta$, the rotation term acts on the weighted sensitivity sum $\sum_i w_i q_i$.

Combining the three contributions, the measurement uncertainty is
\begin{equation}
U_k = \sum_{i \in S_k} w_i^2\, U_i^{\mathrm{stereo}} + U_t
+ \sigma_\theta^2 \left\| \sum_{i \in S_k} w_i\, q_i \right\|^2 .
\label{eq:measurement_uncertainty}
\end{equation}
Because the rotation term couples the weights through $\|\sum_i w_i q_i\|^2$, the weights $w_i$ are obtained by minimizing $U_k$ subject to the simplex constraint in Eq.~(\ref{eq:weighted_measurement}), which admits a closed-form solution refined by a short active-set pass for nonnegativity. Consequently, support features with lower stereo uncertainty and lower attitude sensitivity receive larger weights.

\begin{figure*}[t]
    \centering
    \begin{subfigure}[t]{0.19\textwidth}
        \centering
        \includegraphics[width=\linewidth, height=2.2cm]{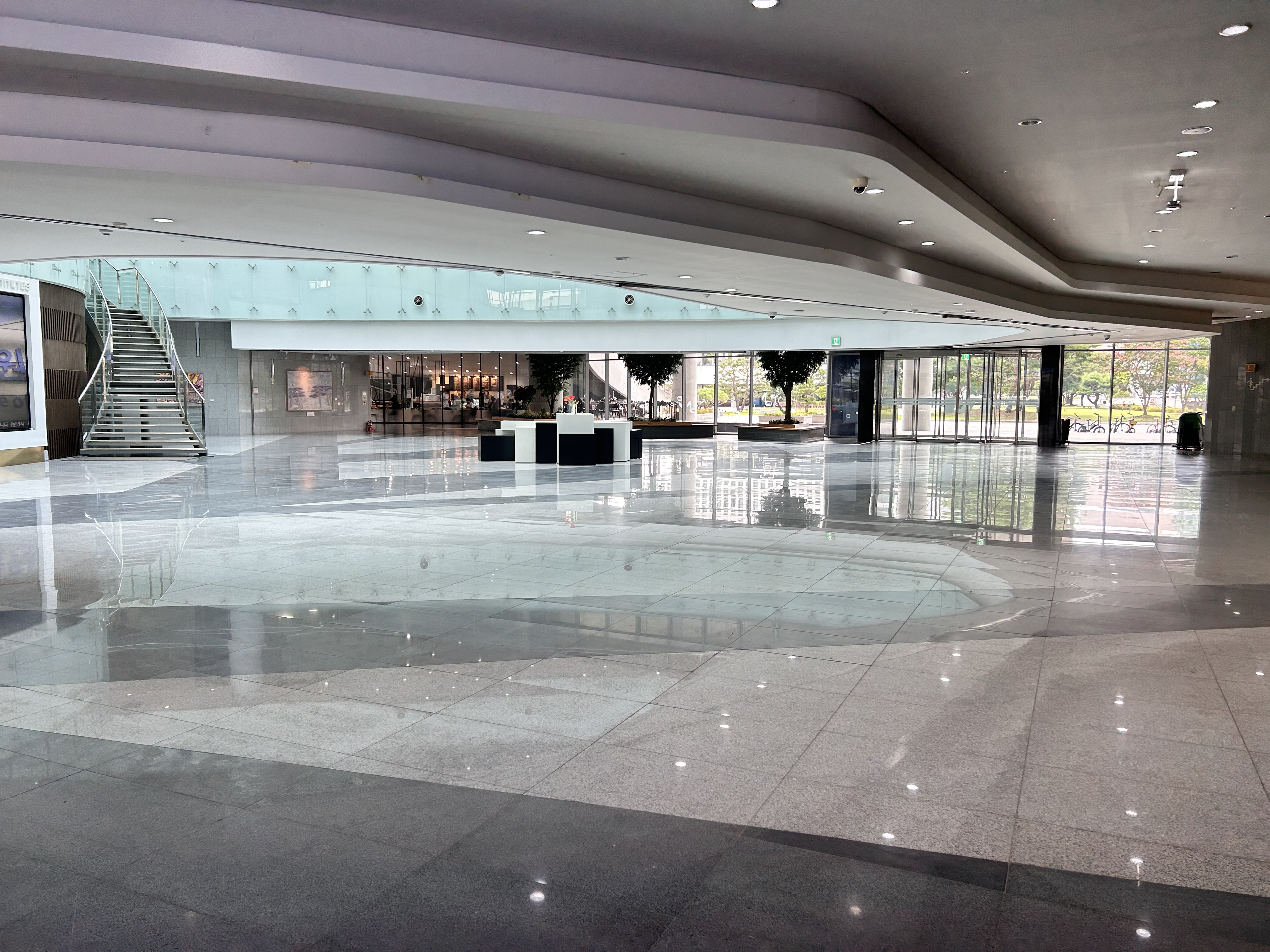}
        \caption{E4}
        \label{fig:env_e4}
    \end{subfigure}
    \hfill
    \begin{subfigure}[t]{0.19\textwidth}
        \centering
        \includegraphics[width=\linewidth, height=2.2cm]{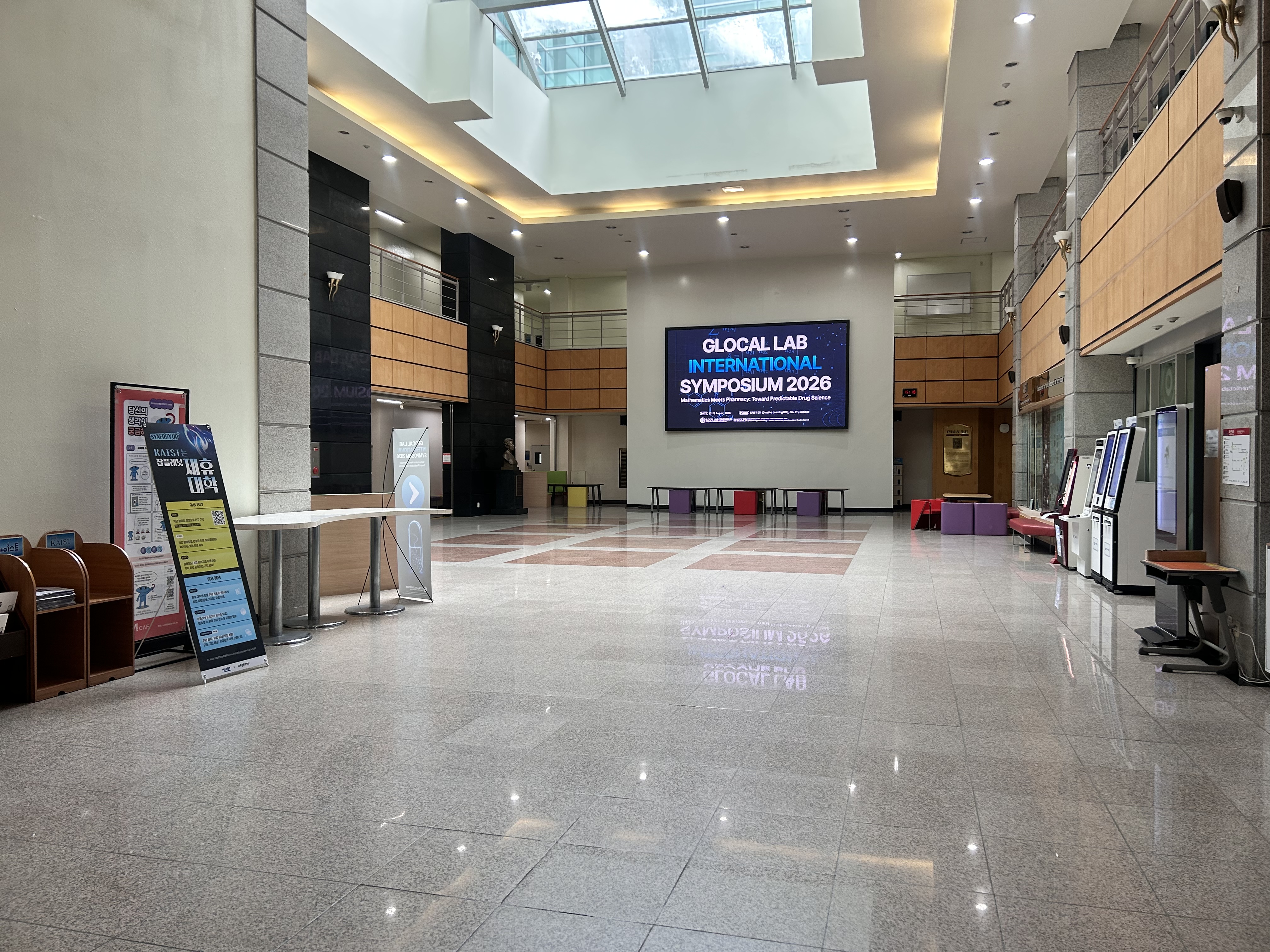}
        \caption{E11}
        \label{fig:env_e11}
    \end{subfigure}
    \hfill
    \begin{subfigure}[t]{0.19\textwidth}
        \centering
        \includegraphics[width=\linewidth, height=2.2cm]{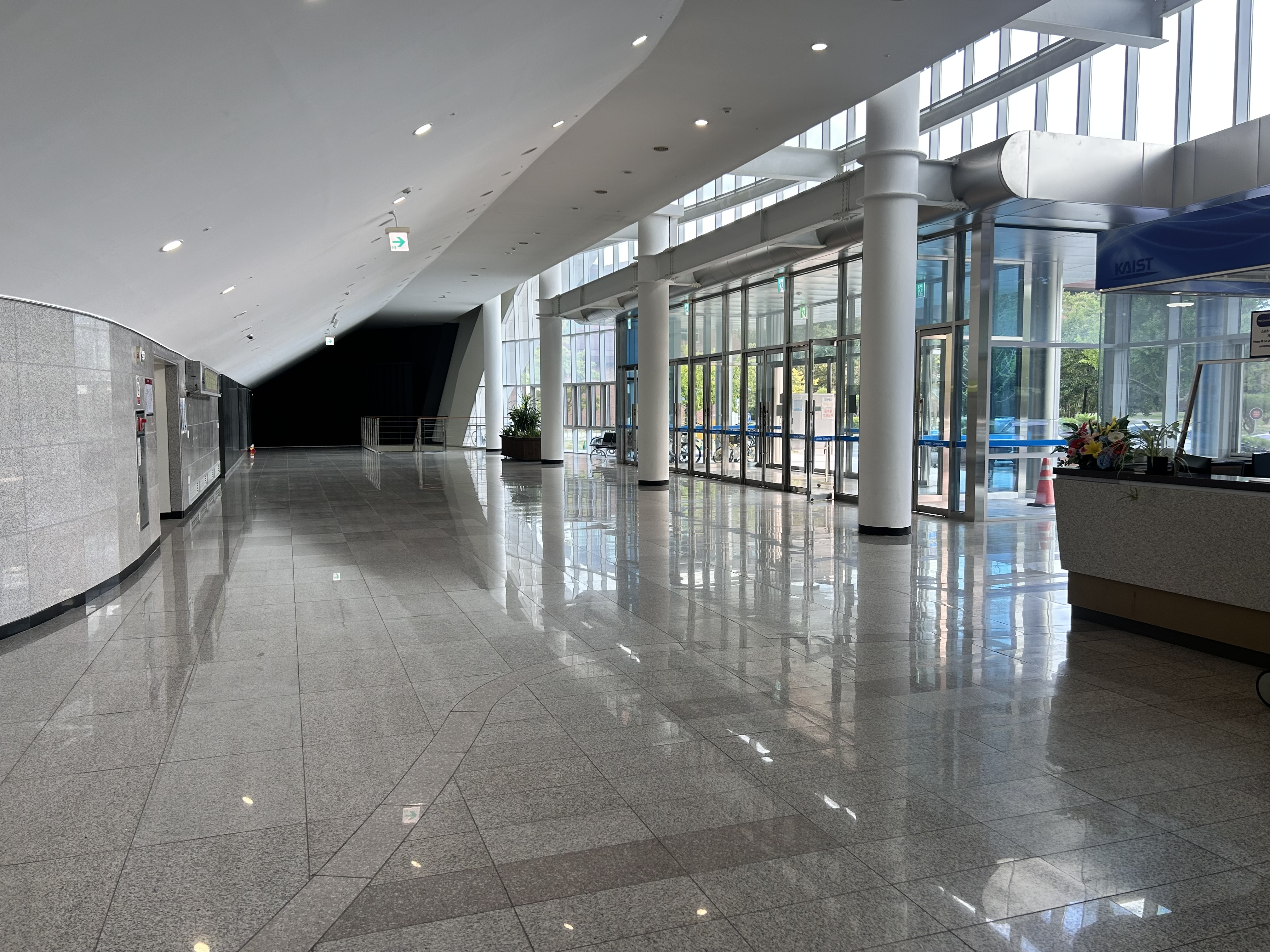}
        \caption{N3}
        \label{fig:env_n3}
    \end{subfigure}
    \hfill
    \begin{subfigure}[t]{0.19\textwidth}
        \centering
        \includegraphics[width=\linewidth, height=2.2cm]{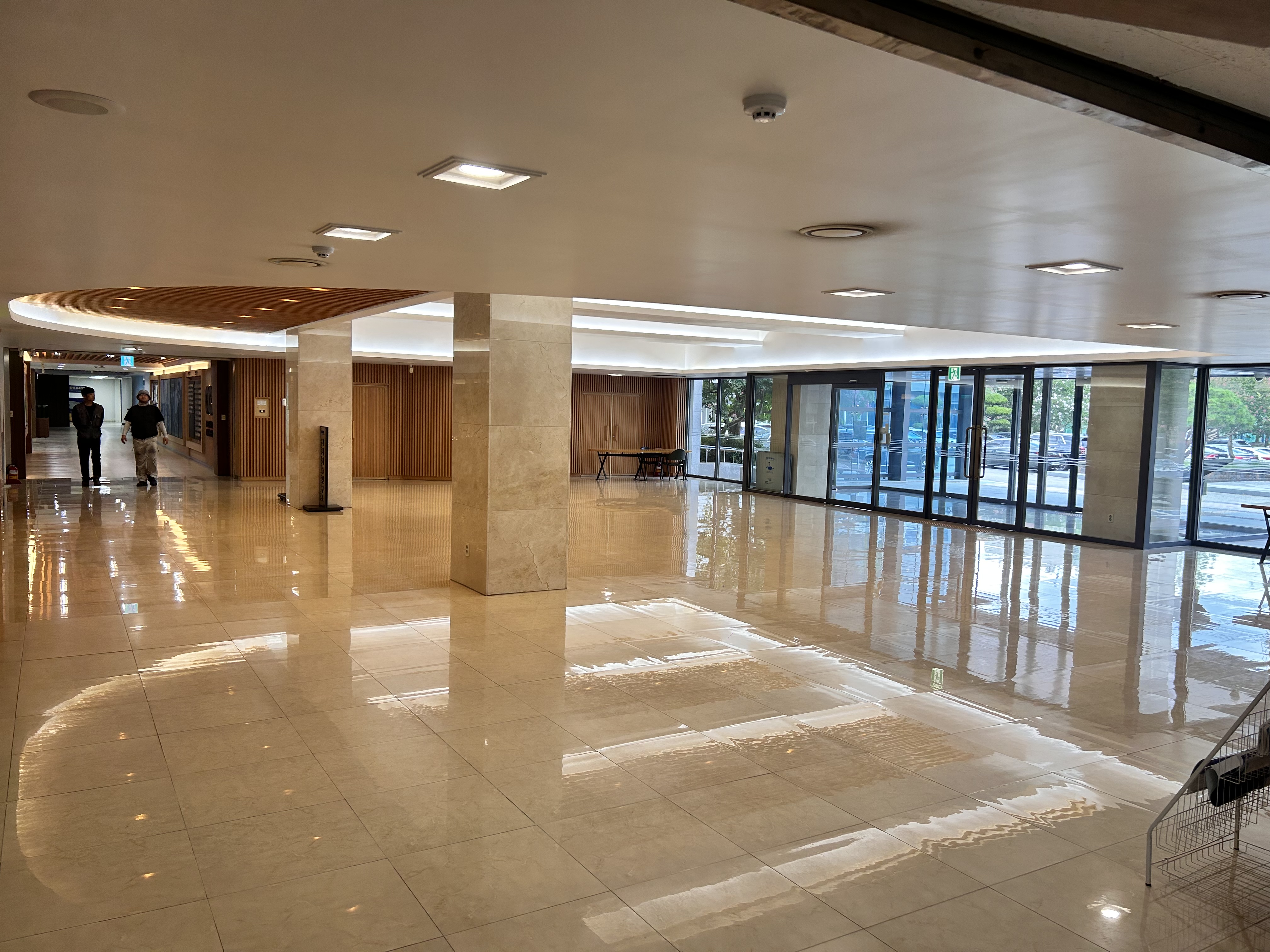}
        \caption{N7}
        \label{fig:env_n7}
    \end{subfigure}
    \hfill
    \begin{subfigure}[t]{0.19\textwidth}
        \centering
        \includegraphics[width=\linewidth, height=2.2cm]{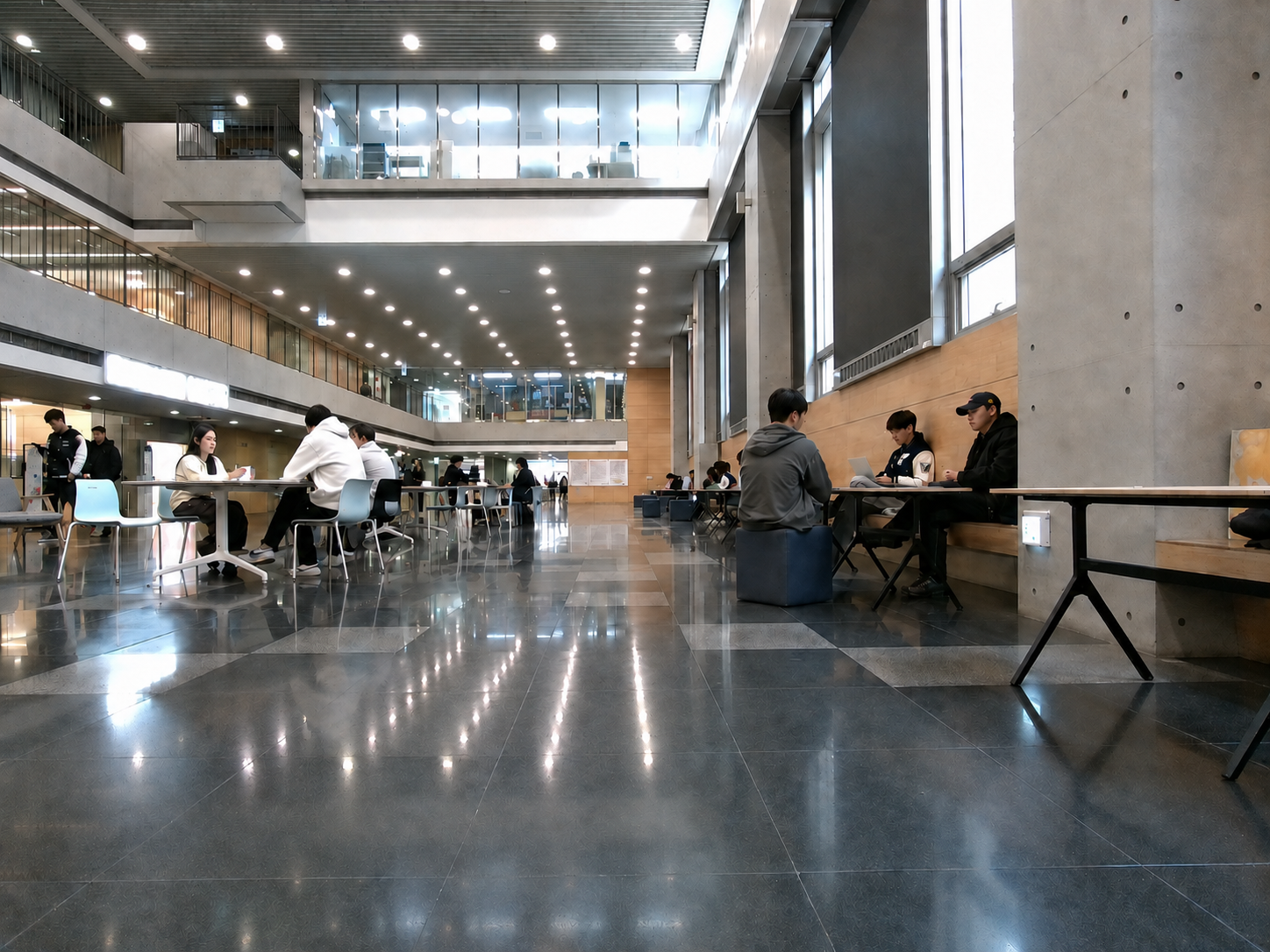}
        \caption{Yonsei}
        \label{fig:env_yonsei}
    \end{subfigure}
    \vspace{-0.8ex}
    \caption{Representative RGB images of the experimental environments: (a) E4, (b) E11, (c) N3, (d) N7, and (e) Yonsei. All environments contain extensive reflective floor regions.}
    \label{fig:experimental_environments}
\end{figure*}

\subsection{Kalman Filter-Based Ground-Height Estimation}
A single per-frame measurement $\tilde{h}_k$ can be noisy or contaminated since $S_k$ could include some above-floor or virtual features. To mitigate this, we introduce a scalar Kalman filter that smooths the ground-height estimate over time, while an NIS gate rejects measurements inconsistent with the current state.

\textbf{State and motion model.} Under the flat indoor-floor assumption, the ground height is modeled as a scalar random-walk state
\begin{equation}
h_k = h_{k-1} + \eta_k,\quad 
\eta_k \sim {N}(0,\,Q_k),
\label{eq:random_walk}
\end{equation}
where the process noise variance $Q_k = \sigma_h^2\,\Delta t_k$ allows slow temporal variation of the ground height.

\textbf{Prediction.} Let $\hat{h}_{k-1}$ and $P_{k-1}$ denote the previous estimate and its error variance. The predicted ground height and prediction-error variance are
\begin{equation}
\hat{h}_k^- = \hat{h}_{k-1},\quad P_k^- = P_{k-1} + Q_k ,
\label{eq:prediction}
\end{equation}
where the superscript $-$ denotes a predicted quantity prior to the current measurement.

\textbf{NIS validation gating.} Before the update, the measurement $\tilde{h}_k$ with uncertainty $U_k$ is tested by the NIS gate~\cite{barshalom2001estimation}, a standard innovation-consistency check:
\begin{equation}
{\mathrm{NIS}}_k = \frac{(\tilde{h}_k - \hat{h}_k^-)^2}{P_k^- + U_k} \leq \gamma_{\mathrm{NIS}} .
\label{eq:nis_gate}
\end{equation}
$\mathrm{NIS}_k$ follows a $\chi^2$ distribution with one degree of freedom, and $\gamma_{\mathrm{NIS}}$ is the corresponding threshold. A measurement violating the gate is discarded and the predicted state is retained, preventing inconsistent measurements from biasing the estimate.

\textbf{Update.} When the measurement passes the gate, the standard scalar Kalman update is applied:
\begin{equation}
K_k = \frac{P_k^-}{P_k^- + U_k},
\label{eq:kalman_gain}
\end{equation}
\begin{equation}
\begin{gathered}
\hat{h}_k = \hat{h}_k^- + K_k(\tilde{h}_k - \hat{h}_k^-),\\
P_k = (1-K_k)P_k^- .
\end{gathered}
\label{eq:kalman_update}
\end{equation}

\textbf{Initialization and recovery.} At initialization, the filter has no stabilized estimate and the NIS gate is not yet effective, so a reliable initial measurement is required. To obtain one, additional candidates are densely extracted inside the floor mask only during the initialization, reducing the risk of inaccurate measurements from sparse floor features. The initial state is then set to the median of five valid per-frame measurements. Once recursive estimation is running, the latest consecutive NIS-rejected measurements are checked for mutual consistency. If their values agree within a fixed band, the filter is reset to their mean. This enables rapid recovery of the estimate when the actual floor height physically changes.

\subsection{Below-Ground Virtual Feature Rejection}
Each floor-region feature in $F_k^{\mathrm{floor}}$ is tested against the estimated ground height $\hat{h}_k$ through the signed height gap
\begin{equation}
e_i = z_i - \hat{h}_k .
\label{eq:height_gap}
\end{equation}
Since $z_i$ increases along the downward gravity direction, a feature located below the estimated ground height satisfies $z_i>\hat{h}_k$, yielding a positive $e_i$.
Whether such a gap is significant depends on how reliable the two heights are. We therefore quantify the per-feature height uncertainty
\begin{equation}
U_i = U_i^{\mathrm{stereo}} + U_t + \sigma_\theta^2 \|q_i\|^2 ,
\label{eq:feature_uncertainty}
\end{equation}
which reuses the stereo, translation, and rotation terms of Section~2.2 but is evaluated per feature rather than aggregated over $S_k$. Since $e_i$ depends on both ground-height and feature-height estimates, its variance is approximated as $\mathrm{Var}(e_i) \approx P_k + U_i$, yielding an uncertainty-aware safety margin
\begin{equation}
\delta_i = \beta\sqrt{P_k + U_i},
\label{eq:safety_margin}
\end{equation}
where $\beta$ controls the confidence width. A floor-region feature is rejected as a below-ground virtual candidate when $e_i > \delta_i$. Because the margin widens when either height is uncertain, valid on-floor features are less likely to be discarded under noisy estimates.

This process allows the final estimator input to preserve the full background set and only the floor-region features satisfying the complementary valid condition:
\begin{equation}
F_k^{\mathrm{valid}}
= F_k^{\mathrm{bg}} \cup
\{f_i \in F_k^{\mathrm{floor}} \mid e_i \leq \delta_i\} .
\label{eq:valid_set}
\end{equation}

\section{Experiments and Results}
\subsection{Experimental Setup}

\begin{table}[!b]
\centering
\caption{Properties of the collected dataset sequences.}
\vspace{-1ex}
\label{tab:sequence_properties}
\scriptsize
\resizebox{\columnwidth}{!}{
\begin{tabular}{lccccc}
\hline
\textbf{Sequence}
& \textbf{XY Area}
& \textbf{Path Length}
& \textbf{Height Change} \\
& \textbf{[m$^2$]}
& \textbf{[m]}
& \\
\hline
E4        & 1499.96 & 184.07 & No \\
E11       & 944.83  & 236.55 & No \\
N3        & 1996.42 & 198.54 & No \\
N7        & 2976.38 & 161.41 & Yes \\
N7 (hand) & 3294.29 & 152.18 & Yes \\
Yonsei 1  & 536.04  & 162.35 & No \\
Yonsei 2  & 728.47  & 207.33 & No \\
Yonsei 3  & 379.35  & 113.40 & No \\
\hline
\end{tabular}
}
\end{table}

To evaluate the proposed method, experiments are conducted from two perspectives: VIO trajectory performance and ground-height estimation accuracy. Since no suitable public stereo-inertial dataset with reflective indoor floors was available, real indoor data were collected in four KAIST buildings (E4, E11, N3, and N7) and in a Yonsei University building, as shown in Fig. 3. The sequence properties are summarized in Table 1, where the height change label indicates whether the environment contains a real ground-height discontinuity, such as a step. A small GCNet~\cite{yang2025gcnet} (GCNet-S) was trained from scratch on real indoor data collected in Yonsei University and KAIST environments together with the Open-LORIS robotic vision dataset~\cite{she2020openloris}, for 20,000 iterations with a batch size of 4 using SGD. No frames from the evaluation sequences were used for training; the collected data were separated at the building level.

\begin{table*}[t]
\centering
\caption{ATE RMSE [m] / std [m] comparison on reflective-floor indoor sequences. The reported values are the medians over five runs. The smallest ATE RMSE and std values are shown in bold, and divergent cases are denoted by ``--''.}
\vspace{-1ex}
\label{tab:ate_comparison}
\scriptsize
\resizebox{\textwidth}{!}{
\begin{tabular}{lcccccccc}
\hline
\textbf{Method}
& \textbf{E4}
& \textbf{E11}
& \textbf{N3}
& \textbf{N7}
& \textbf{N7 (hand)}
& \textbf{Yonsei 1}
& \textbf{Yonsei 2}
& \textbf{Yonsei 3} \\
\hline
VINS-Fusion
& 0.722/0.405& \textbf{1.238}/\textbf{0.364}& 0.845/0.558
& \textbf{0.265}/\textbf{0.125}
& 0.538/0.224
& 1.394/0.726
& 1.416/0.577
& 0.853/\textbf{0.473} \\

VINS-Fusion (reject floor)
& 0.592/0.313
& 1.365/0.444
& 0.762/0.366
& 0.510/0.229
& 0.520/0.247
& 1.564/0.915
& 1.313/0.528
& 0.894/0.620 \\

VINS-Fusion + Ours
& \textbf{0.335}/\textbf{0.115}
& 1.317/0.424
& \textbf{0.460}/\textbf{0.215}
& 0.338/0.138
& \textbf{0.501/0.222}
& \textbf{1.168/0.660}
& \textbf{1.303/0.511}
& \textbf{0.742}/0.503 \\
\hline
DynaVINS
& --
& --
& 2.557/0.685
& 2.147/1.109
& --
& --
& 1.335/0.705
& 0.987/0.405 \\

DynaVINS (reject floor)
& 0.693/0.387
& 0.762/0.333
& 0.809/0.393
& 0.497/0.107
& --
& 1.232/0.674
& 1.433/0.706
& 0.932/0.463 \\

DynaVINS + Ours
& \textbf{0.537/0.328}
& \textbf{0.759/0.289}
& \textbf{0.550/0.348}
& \textbf{0.351/0.102}
& \textbf{0.625/0.250}
& \textbf{1.168/0.632}
& \textbf{1.227/0.511}
& \textbf{0.615/0.340} \\
\hline
\end{tabular}
}
\end{table*}

Data were acquired using an Intel RealSense D455 stereo camera, an Xsens MTI-100 IMU, and a Livox Mid-360 LiDAR. Except for the N7 (hand) sequence, the data were collected with the sensor rig mounted on a cart. The N7 (hand) sequence was collected by hand with deliberate translational and rotational motion to examine the method's stability under such pose variation. All experiments were run on an Intel Core i9-13900HX CPU and an NVIDIA RTX 4080 Laptop GPU, with filter parameters set to $N=3$, $\gamma_{\mathrm{NIS}}=10$, $\beta=3$,  $\sigma_\theta=0.1^\circ$, $\sigma_t=0.01$~m, $P_0=0.003$~m$^2$, and $\sigma_h^2=3\times10^{-4}$~m$^2$/s.
\begin{figure}[!t]
\centering
\includegraphics[width=0.99\columnwidth]{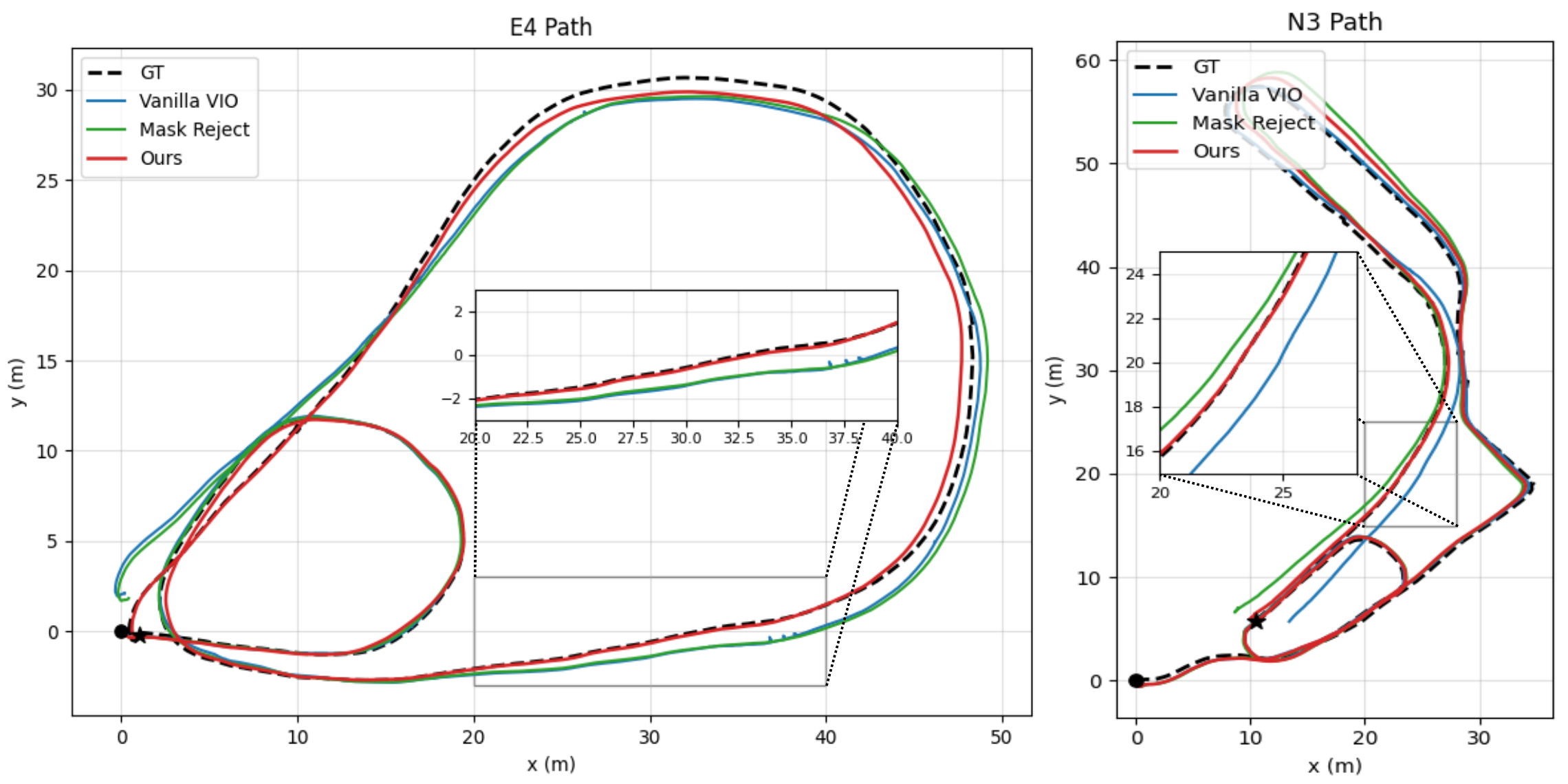}
\vspace{-3ex}
\caption{VINS-Fusion trajectories on the E4 and N3 sequences (original, naive floor rejection, and ours vs.\ the FAST-LIO2 reference).}
\label{fig:results_path}
\end{figure}

For trajectory evaluation, the proposed front-end filter is applied to two representative feature-based VIO front-ends, VINS-Fusion~\cite{vinsmono,qin2019a} and DynaVINS~\cite{dynavins}; the latter is included to discount the dynamic-object effects present in the collected data sequences. On each front-end, we compare three configurations—original VIO, naive floor rejection, and the proposed method. Accuracy is measured by the Absolute Trajectory Error (ATE) against a FAST-LIO2~\cite{fastlio2} reference trajectory, reported as Root Mean Square Error (RMSE) and standard deviation (std) after aligning each estimate to the reference.

For ground-height evaluation, an independent camera-to-ground reference is constructed offline from synchronized Livox LiDAR measurements. A local ground plane is fitted using RANSAC and transformed into the camera frame using calibrated LiDAR-to-camera extrinsics. The distance from the camera center to the plane along the gravity direction is used as the reference. For our estimate, the corresponding distance is obtained from $\hat{h}_k$ and the camera position along gravity. The per-frame absolute error is computed between the two distances.

\subsection{Experimental Results}
\subsubsection{VIO Trajectory Performance}

Table~\ref{tab:ate_comparison} shows that the proposed method improves localization accuracy on six of eight sequences with VINS-Fusion~\cite{vinsmono,qin2019a}. With DynaVINS~\cite{dynavins}, it restores convergence on the four sequences where the baseline diverges and reduces the error on the remaining sequences. It also outperforms naive floor rejection on all converged sequences, demonstrating the benefit of preserving valid on-floor features. The representative trajectories in Fig.~\ref{fig:results_path} show closer agreement with the FAST-LIO2 reference. The slight degradation on N7 may result from its highly specular floor, where mirror-like virtual features can provide useful constraints.

\subsubsection{Ground-Height Estimation Accuracy}

Table~\ref{tab:ground_height_accuracy} reports the per-frame absolute ground-height errors against the independent LiDAR-derived reference. Except for E11, the mean error remains at or below 0.05~m, indicating stable ground-height estimation. E11 shows a larger mean error of 0.088~m and std of 0.127~m mainly due to dark corridor scenes, where floor-region masking is less reliable and floor features are often sparse. The filter therefore relies more on prediction, which can lead to error accumulation and excessive feature removal. This also explains the slight trajectory degradation in E11 compared with the original VIO.

\begin{table}[!t]
\centering
\caption{Ground-height absolute-error statistics on sequences without actual ground-height changes. Mean, std, and percentiles are computed over the per-frame absolute errors against the LiDAR-derived reference.}
\vspace{-1ex}
\label{tab:ground_height_accuracy}
\scriptsize
\resizebox{\columnwidth}{!}{
\begin{tabular}{lcccccc}
\hline
\textbf{Sequence}
& \textbf{Mean}
& \textbf{Std}
& \textbf{30\%}
& \textbf{50\%}
& \textbf{90\%} \\
& \textbf{[m]} & \textbf{[m]} & \textbf{[m]} & \textbf{[m]} & \textbf{[m]} \\
\hline
E4       & 0.029 & 0.044 & 0.009 & 0.018 & 0.066 \\
E11      & 0.088 & 0.127 & 0.023 & 0.064 & 0.207 \\
N3       & 0.050 & 0.049 & 0.018 & 0.032 & 0.098 \\
Yonsei 1 & 0.044 & 0.052 & 0.018 & 0.029 & 0.085 \\
Yonsei 2 & 0.038 & 0.048 & 0.014 & 0.026 & 0.079 \\
Yonsei 3 & 0.049 & 0.057 & 0.021 & 0.038 & 0.099 \\
\hline
\end{tabular}
}
\end{table}

\begin{figure}[!b]
\centering
\includegraphics[width=\columnwidth]{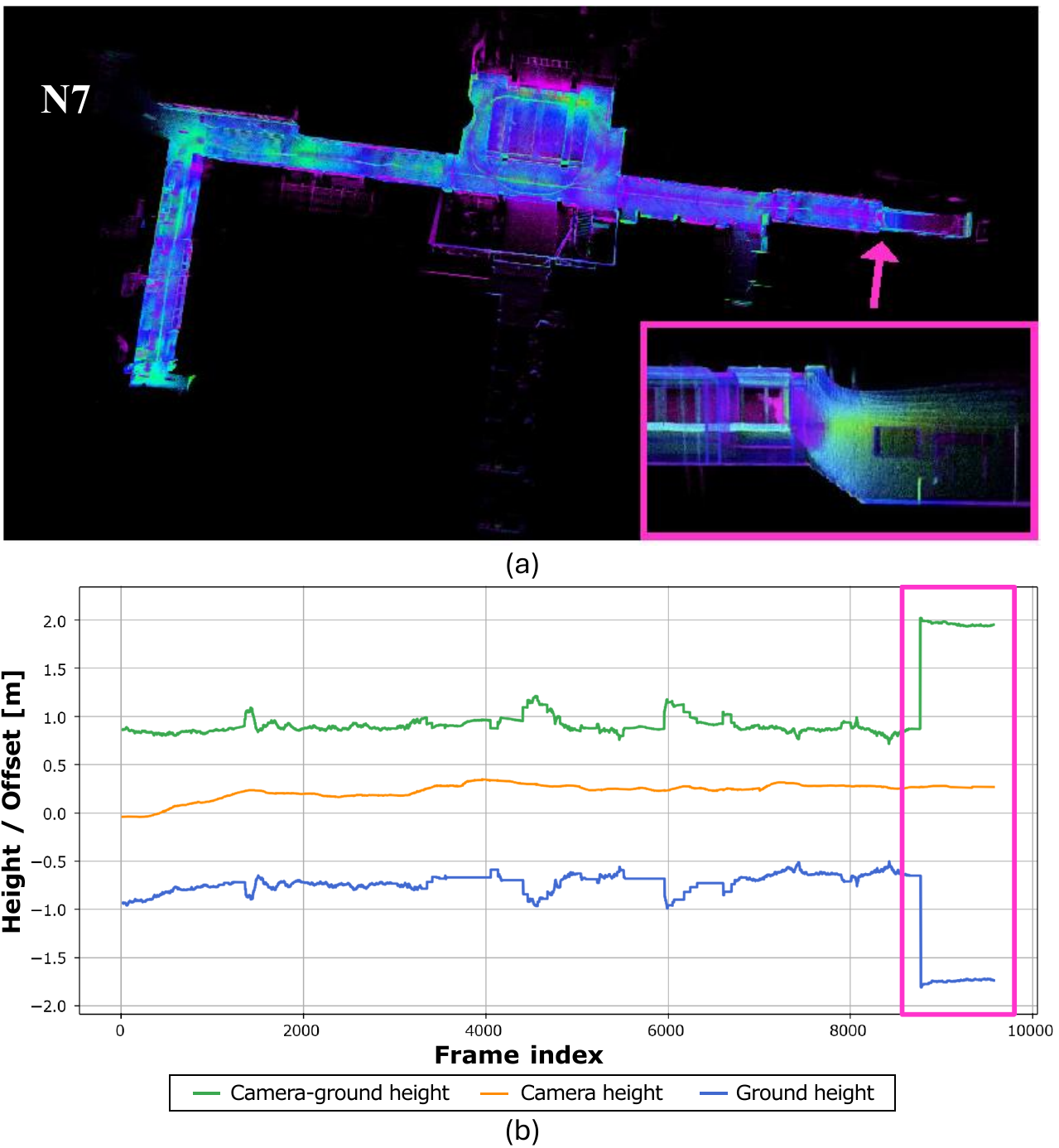}
\caption{Ground-height estimation results in the N7 sequence. (a) LiDAR map and (b) per-frame height estimates, with the ground-height change region highlighted.}
\label{fig:n7_ground_height}
\end{figure}

The proposed method also handles real ground-height changes. Fig.~\ref{fig:n7_ground_height} shows the recovery behavior in the N7 sequence, where a descending stair region changes the physical ground height. As shown in Fig. 5 (b), the estimated camera-to-ground distance remains stable before the highlighted region. When the lower stair floor enters the camera view, the incoming measurements become inconsistent with the previous estimate and are initially rejected by the NIS gate. However, once the rejected measurements become mutually consistent, the filter resets to the new ground height. This demonstrates that the estimator can reject transient corrupted measurements while still adapting to actual floor-height changes.

\subsubsection{Runtime}
GCNet-S runs asynchronously in a separate node and requires $8.96$~ms per image on average, sustaining approximately $30$~Hz mask output for the $30$~Hz image stream. The proposed geometric filtering requires $4.93$~ms per VIO update. With the proposed front-end, the measured VINS-Fusion processing rate decreases only slightly from $15.0$~Hz to $14.7$~Hz, demonstrating that the additional computation preserves real-time operation.

\section{CONCLUSION} This paper presented a reflection-robust front-end that estimates ground height from sparse stereo features and selectively rejects below-ground virtual features using uncertainty-aware Kalman filtering. Experiments with VINS-Fusion and DynaVINS demonstrate improved robustness on most reflective-floor sequences while maintaining real-time operation. Future work will investigate robust segmentation in weakly textured scenes and selective use of stable mirror-like virtual features.

\section*{ACKNOWLEDGEMENT}
This work was supported by the Korea Evaluation Institute Of Industrial Technology (KEIT) grant funded by the Korea government (MOTIE) (No.20023455, Development of Cooperate Mapping, Environment Recognition and Autonomous Driving Technology for Multi Mobile Robots Operating in Large-scale Indoor Workspace). The students are supported by BK21 FOUR.


\end{document}